\pdfoutput=1

\documentclass[11pt]{article}

\usepackage[final]{acl}

\usepackage{times}
\usepackage{latexsym}

\usepackage[T1]{fontenc}
\usepackage{bbm}
\usepackage{tabularx}
\usepackage{amsmath}
\usepackage{amssymb}
\usepackage{algorithm}
\usepackage{algpseudocode}
\usepackage{subcaption}
\usepackage{multirow}
\usepackage{booktabs} 
\usepackage{adjustbox}
\usepackage{tabularx}
\usepackage{graphicx} 
\usepackage{enumitem}
\usepackage[T1]{fontenc}
\usepackage{aecompl}
\usepackage{float}
\usepackage{balance}
\usepackage[utf8]{inputenc}

\usepackage{microtype}

\usepackage{inconsolata}

\usepackage{graphicx}

\title{P3: A Policy-Driven, Pace-Adaptive, and Diversity-Promoted Framework for data pruning in LLM Training}

\newcommand{\customsize}{\fontsize{10.5pt}{12pt}\selectfont}

\author{
    Yingxuan Yang\textsuperscript{1}, Huayi Wang\textsuperscript{1}, Muning Wen\textsuperscript{1}, Xiaoyun Mo\textsuperscript{2}, Qiuying Peng\textsuperscript{2}\\ \textbf{Jun Wang}\textsuperscript{2},\textbf{Weinan Zhang}\textsuperscript{1} \\
    \textsuperscript{1}\customsize Shanghai Jiao Tong University, Shanghai, China \\
    \textsuperscript{2}\customsize Oppo Research Institute, Shenzhen, China \\
    \customsize \texttt{\{zoeyyx, wnzhang\}@sjtu.edu.cn}
}

\begin{document}
\maketitle
\begin{abstract}
In the rapidly advancing field of Large Language Models (LLMs), effectively leveraging existing datasets during fine-tuning to maximize the model’s potential is of paramount importance. This paper introduces P3, an adaptive framework aimed at optimizing the task-specific fine-tuning process through iterative data pruning. \textbf{P3} consists of three key components: (1) \textbf{\underline{P}}olicy-driven Difficulty Measurement, which dynamically assesses data difficulty based on the model's real-time performance, replacing static metrics with adaptable evaluations; (2) \textbf{\underline{P}}ace-Adaptive Selection, leveraging self-paced learning to progressively introduce more challenging data, thereby enhancing model capability; (3) Diversity \textbf{\underline{P}}romotion, incorporating Determinantal Point Process (DPP) to ensure data diversity across epochs, enriching the learning process. We validate \textbf{P3} on the reasoning scenarios, APPS and MATH, demonstrating significant improvements over traditional data pruning methods. By advancing dynamic data selection and utilization strategies, \textbf{P3} contributes both a theoretical framework and concrete approach to fully exploit existing data for LLMs' performance improvement, offering utility across diverse tasks.
\end{abstract}

\section{Introduction}
As Large Language Models (LLMs) continue to evolve rapidly, their ability to handle specialized tasks has become a key focus of research in both academia and industry."\citep{brown2020languagemodelsfewshotlearners, mishra2022crosstaskgeneralizationnaturallanguage, touvron2023llama2openfoundation, openai2024gpt4technicalreport}. This trend underscores the growing importance of LLMs in advancing various technological applications \citep{naveed2024comprehensiveoverviewlargelanguage, ai4science2023impactlargelanguagemodels, zhou2024surveylargelanguagemodels}. Deploying LLMs within specific domains normally requires a process called fine-tuning, which tailors the model’s capabilities to meet the particular needs of different applications. Fine-tuning typically involves adjusting the models using training data that closely reflects specific (application) scenarios \citep{Dong2023HowAI, alpaca}.

\begin{figure}[t]
    \centering
    \includegraphics[width=\linewidth]{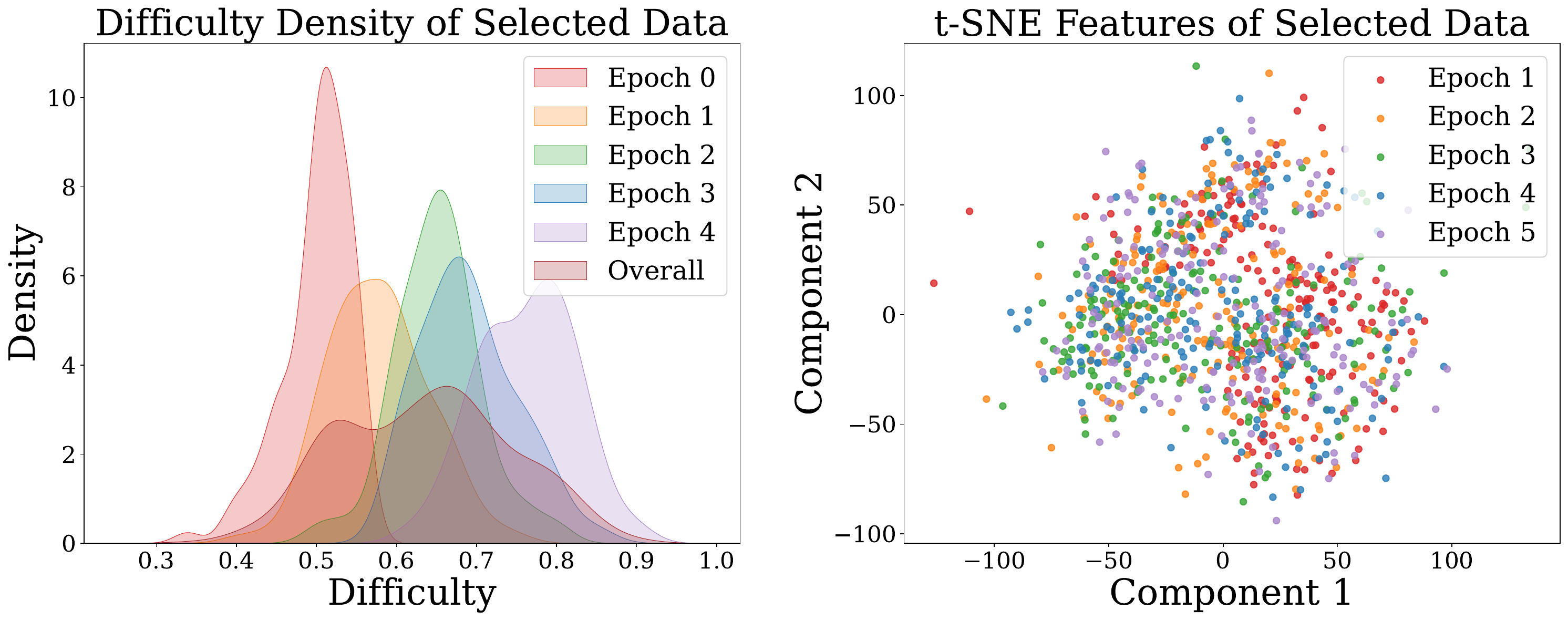}
    \vspace{-0.3cm}
    \caption{Training Data Analysis Across Epochs. The left graph shows the difficulty distribution of data selected in different epochs, illustrating our progressive training strategy from easier to more challenging tasks. The right graph, using t-SNE visualization, displays the uniform and diverse distribution of selected data.}
    \label{fig:analysis}
    \vspace{-10pt} 
\end{figure}

In the supervised fine-tuning (SFT) of large language models (LLMs), selecting data before or during training is crucial. Research \citep{albalak2024surveydataselectionlanguage, touvron2023llama2openfoundation, zhou2024lima, li2024quantityqualityboostingllm, hase2024unreasonableeffectivenesseasytraining, li2024superfiltering} shows that LLMs can perform as well as, or even better than, models trained on large datasets by using fewer but higher-quality data points. Both the LLaMA2 and LIMA studies emphasize that "\textbf{Quality Is All You Need}" and that "\textbf{Less is more for alignment}", demonstrating that a smaller set of well-curated data can yield significant results in SFT of LLMs \citep{touvron2023llama2openfoundation, zhou2024lima}. These findings consistently highlight that the success of fine-tuning LLMs heavily relies on data quality.

Current data pruning methods, including manually defined rules and scoring models to define or evaluate the quality \citep{li2024quantityqualityboostingllm, hase2024unreasonableeffectivenesseasytraining, li2024superfiltering}, have notable limitations: (1) They rely on static datasets, failing to adapt to the evolving requirements at different training stages, which often leads to overfitting; (2) Manually defined heuristics and scoring models are not universally applicable, resulting in inaccurate data valuation when applied across different models or datasets. Furthermore, current data selection methods are usually predetermined before training, however evidence~\cite{NEURIPS2023_b9e472cd} suggests that LLMs tend to overfit when trained on static, uniform datasets. 

Therefore, our goal is to develop a dynamic and iterative data selection framework during training for LLM that adapts to the changing requirements of each training phase. Previous research \citep{swayamdipta-etal-2020-dataset, azeemi-etal-2023-data} has shown that training with appropriately challenging data can improve model performance, highlighting the need for adaptive difficulty. Leveraging self-paced learning, which gradually increases data complexity, allows us to optimize the learning process efficiently, as demonstrated in other domains like semi-supervised learning and noisy label handling. Additionally, maintaining data diversity is essential to promote generalization and reduce overfitting throughout training \citep{albalak2024surveydataselectionlanguage}.

Motivated by these insights, we propose a novel adaptive data pruning framework, \textbf{P3}, to enhance LLM fine-tuning. As shown in Figure~\ref{fig:analysis}, our approach dynamically subsets of training data tailored to different stages of the model's lifecycle, emphasizing higher precision on the same dataset. Specifically, \textbf{P3} integrates policy-driven difficulty assessment, self-paced learning (SPL), and Determinantal Point Process (DPP) to progressively fine-tune the model, moving from easier to more challenging tasks while maintaining data diversity.
Our key contributions include:
\vspace{-0.2cm}
\begin{itemize}[itemsep=1.5pt]
    \item \textbf{Policy-Driven Difficulty Measurement}: A policy-based data evaluation method that directly uses model-derived policies to assess data utility, providing a more accurate and objective alignment with the model's real-time performance.
    \item \textbf{Pace-Adaptive selection for Training}: A stage-wise training strategy using self-paced learning to improve precision with fewer data points. This represents one of the first uses of self-paced learning to optimize LLM fine-tuning for task-specific objectives.
    \item \textbf{Diversity-Promoted Selection via DPP}: An innovative integration of DPP with self-paced learning to balance training data \textbf{difficulty} and \textbf{diversity}, achieving competitive results
 in reasoning tasks for LLMs.
\end{itemize}

Our experiments on the APPS and MATH datasets show that \textbf{P3} significantly outperforms traditional methods, demonstrating the effectiveness of our dynamic training strategies in enhancing model performance.

\section{Related Works}
\subsection{Data Pruning}
Data pruning, or data selection, has been widely studied to enhance model performance while reducing training dataset size. The goal is to select a subset of data that enables the model to achieve optimal performance~\citep{maharana2023d2, pmlr-v162-mindermann22a}. Numerous coreset selection techniques have been developed to score data points, indicating their quality for training, with higher scores suggesting greater relevance. 
However, there is still limited research focused specifically on the application of these methods in the domain of LLMs~\citep{albalak2024surveydataselectionlanguage,li2024quantityqualityboostingllm,hase2024unreasonableeffectivenesseasytraining}. These scoring methods can be categorized into three types: scores based on human expert standards, model-based scores, and scores computed using neural network calculations~\citep{albalak2024surveydataselectionlanguage}.

Our approach differs from previous work in several key aspects. First, we contend that relying on a fixed standard or another model to prioritize data does not ensure that the selected data will be advantageous for training. Additionally, the difficulty measured by one model might not match the difficulty faced by the training model. Furthermore, we have observed that the difficulty of the same data may decrease as training progresses. This indicates that the data for the training model can vary at different stages, rendering scores assigned prior to training less reliable over time. To tackle these issues, we evaluate the model's actual problem-solving capabilities based on the policies generated during various training stages. A lower success rate at a specific stage suggests that the data presents greater challenges for the model at that point.
\vspace{-0.2cm}

\subsection{Adaptive Training Strategy}

Self-paced learning (SPL) \citep{NIPS2010_e57c6b95,jiang2015self,NIPS2014_c60d060b,kong2021adaptive} is well-suited for designing training processes using data of varying difficulties, following a path from simple to complex tasks. SPL allows models to select tasks based on their current learning state, promoting self-regulation. Our research emphasizes dynamic training, using self-paced learning to tailor our approach. Our framework prioritizes the model's autonomy and adaptability, allowing it to select learning content aligned with its current capabilities, thereby facilitating an adaptive learning pace. We also develop a pioneering DPP-based SPL framework that strategically balances the \textbf{difficulty} of tasks with the \textbf{diversity} of training data. We believe applying SPL to optimize task-specific model fine-tuning is a pioneering effort in this field.

\begin{figure*}[t]
    \centering
    \includegraphics[width=0.95\linewidth]{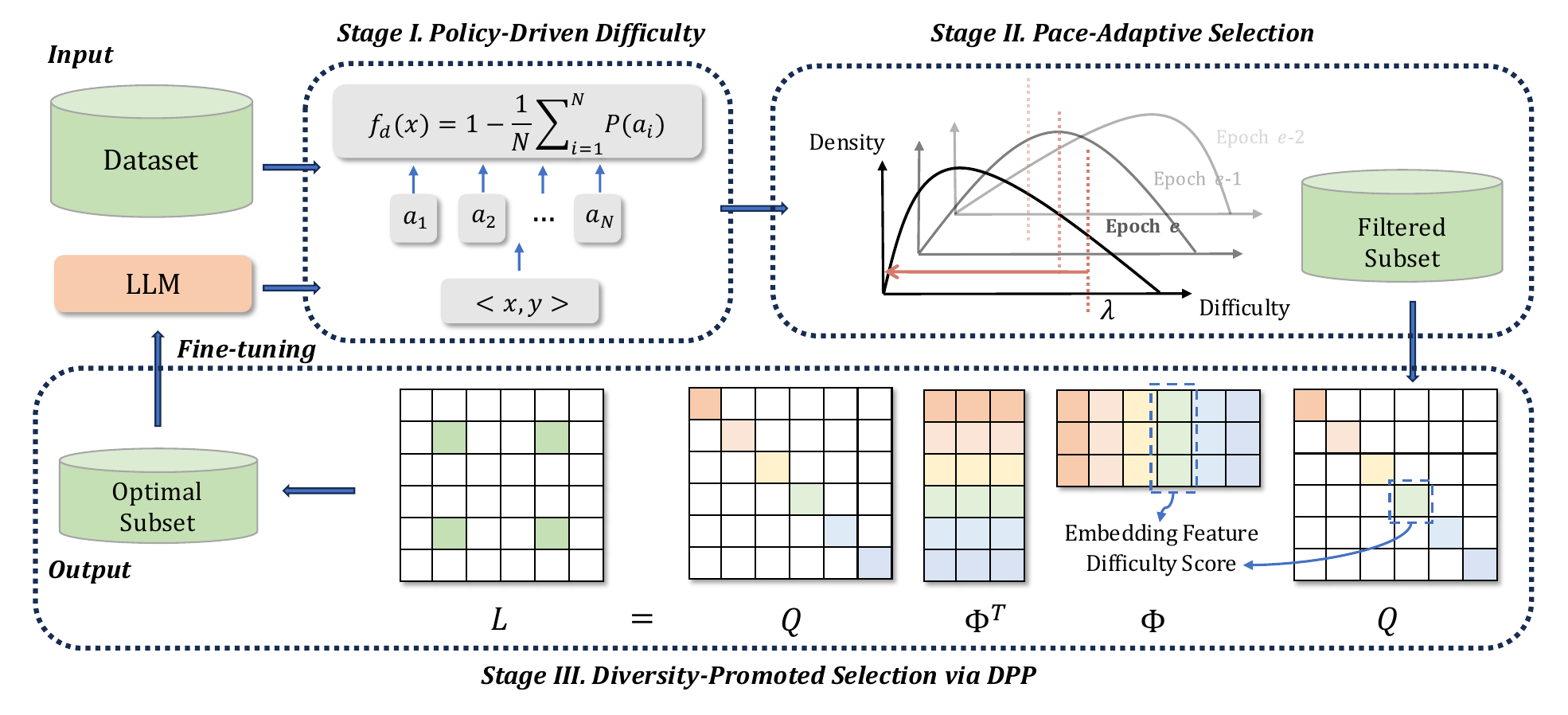}
    \vspace{-0.3cm}
    \caption{Illustration of the P3 Framework, which embodies a Policy-driven, Pace-adaptive, and Diversity-Promoted approach to optimize Large Language Model training through three strategic stages.}
    \label{fig:framework3}
\end{figure*}

\section{Preliminaries}

\subsection{Problem Formulation}
Consider a domain-specific dataset \(D\) consisting of \(N\) samples, denoted as \(\{(x_i, y_i)\}_{i=1}^N\). We utilize a pre-trained model, parameterized by \(\theta\), which will be trained across \(E\) epochs on the dataset.

The primary objective is to iteratively minimize the expected loss on \(D\) by dynamically fine-tuning the pre-trained model \(\theta\) over the training epochs. This process involves strategically selecting an optimal subset \(D_e\) of the training data at the epoch e, determined by the pruning rate \(\alpha_e\):
\begin{equation}
    \min_{D_e \subseteq D: \frac{|D_e|}{|D|} \leq \alpha_e} \mathbb{E}_{x,y} \left[ L(x, y; \theta_e(D_{e-1})) \right].
\end{equation}
Here, \(L\) represents the loss function evaluating the model’s performance on \(D_e\).




\section{Methodology}
As shown in Figure~\ref{fig:framework3}, our \textbf{P3} framework consists of policy-driven difficulty assessment, pace-adaptive training via self-paced learning (SPL), and diversity promotion using Determinantal Point Process (DPP).
\textbf{Section 4.1} details the difficulty assessment, while \textbf{Sections 4.2} and \textbf{4.3} outline SPL and DPP. \textbf{Section 4.4} describes their integration to optimize data pruning, enhance learning efficiency, and improve generalization through balanced quality and diversity.

\subsection{Policy-Driven Difficulty Measurement}
\label{sec:difficulty}

This section introduces a policy-driven difficulty metric that evaluates task complexity by analyzing the decisions made by a LLM during execution. 

In this work, we use "policy" in an RL-inspired manner, though we do not implement reinforcement learning. Here, "policy" refers to the model's implicit decision-making during training, specifically how it assigns probabilities to different generated sequences, which reflects its assessment of task complexity. Unlike perplexity, this policy-driven difficulty measure captures the model's evolving feedback on task complexity, helping to guide data selection effectively.

To assess the challenge of generating outputs for an instruction \(x_i\), we first segment the corresponding output \(y_i\) into discrete sequences \(a_i\):
\begin{equation}
    \{a_1, a_2, \ldots, a_n\} = \text{split}(y_i).
\end{equation}

This segmentation step allows us to analyze complexity at a finer level, identifying which parts of the output contribute most to the task's overall difficulty. It is applicable to various tasks, such as dividing code into lines or breaking down mathematical solutions step-by-step. In essence, generating text can also be viewed as a step-by-step process, making it natural to treat each segment as an action taken by the model.

Next, we compute the generation probability \(P(t_i)\) for each sequence using the model's softmax output. For each sequence, we extract the logits corresponding to the label tokens after the prompt length. The generation probability is defined as:

\[
P(t_i) = \frac{\exp(\text{logit}_{t_i})}{\sum_{j} \exp(\text{logit}_{t_j})},
\]

where \(\text{logit}_{t_i}\) represents the logits for the specific token \(t_i\), and the denominator sums the exponentials of all logits for the tokens in the vocabulary. To ensure comparability across different sequences, these probabilities are normalized. 

The task difficulty for each sample i at the current epoch
 \(e\) is defined as the complement of the average generation probabilities:
 \vspace{-0.2cm}
\begin{equation}
    d_e(x_i, y_i, \theta) = 1 - \frac{1}{n} \sum_{i=1}^{n} P(a_i),
\end{equation}
where \(n\) is the number of sequences. A higher difficulty score indicates a more challenging task for the model.

As the model trains, the difficulty score may change, making initially hard tasks easier over time. This flexibility enables dynamic training data allocation based on the model's current capabilities.

\subsection{Pace-Adaptive Training Methodology}
In this section, we introduce a pace-adaptive data pruning method during training aimed at optimizing the training process for LLMs. We implement self-paced learning (SPL) to select training samples that align with the model's capabilities, ensuring it encounters suitable challenges at each stage to maximize learning efficiency. 

We train the model on a domain-specific dataset \(D = \{(x_i, y_i), i \in [1, N]\}\). During training, we assign a weight \(\nu\) to each data example, where \(\nu = [\nu_1, \nu_2, \ldots, \nu_N] \in [0, 1]^N\). This weight is adjusted according to the difficulty of the samples, which we quantify using a function \(d_e(x_i, y_i, \theta)\) that measures the prediction challenge for each data point.

As training progresses, we iteratively update the difficulty assessment for each example based on the model's policy. This assessment accounts for the model's learning progress, ensuring that the selected training samples reflect the model's current ability. Specifically, we define the adjusted difficulty as:
\begin{equation}
d_{\text{adj}, e}(x_i, y_i, \theta) = d_e(x_i, y_i, \theta) + r(\theta)
\end{equation}

where \(r(\theta)\) is a regularization term designed to prevent overfitting:
\begin{equation}
    r(\theta) = \alpha \left(f_d(x_i, y_i, \theta_{e-1}) - f_d(x_i, y_i, \theta) \right).
\end{equation}

In each training epoch, we employ a self-paced function \(f(\lambda, \nu_i) = -\lambda \nu_i\) to dynamically focus training efforts. This function determines which samples should be emphasized based on the current difficulty and training progress. We introduce a threshold parameter \(\lambda\) to control the rate at which new samples are included. 
\begin{equation}
    \nu^*_{i}(\lambda, l) = 
    \begin{cases} 
    1, & \text{if } d_e(x_i, y_i, \theta) < \lambda \\
    0, & \text{otherwise}
    \end{cases}.
\end{equation}

As training progresses, we define \(\lambda\) as a linear function of the current epoch, specifically a linear function with a slope of 1, increasing from a lower percentile (e.g., 50\%) to a higher percentile (e.g., 95\%). This approach gradually introducing more challenging samples to encourage continual improvement in the model.

\subsection{Diversity-Promoted Selection via DPP}

To ensure that our selected training samples are diverse, we use Determinantal Point Processes (DPPs)~\citep{Kulesza_2012}. DPP helps us select subsets that are both diverse and representative, which is crucial for training an effective model.

\paragraph{Kernel Matrix for DPP}
A core part of DPP is the kernel matrix \(L\), which integrates both the quality and similarity of samples. 

We start by defining a similarity matrix \(S\) that captures how similar each sample is to the others:
\vspace{-0.2cm}
\begin{equation}
    S_{ij} = \frac{\phi_i^\top \phi_j}{\|\phi_i\| \|\phi_j\|},
\end{equation}

where \(\phi_i\) represents the feature vector for sample \(i\). To also consider sample quality, we construct the kernel matrix \(L\) as:
\vspace{-0.2cm}
\begin{equation}
    L_{ij} = q_i S_{ij} q_j,
\end{equation}

where \(q_i\) is the quality score of sample \(i\), which we define as the difficulty score \(d_e(x_i, y_i, \theta)\) from Section~\ref{sec:difficulty}. The feature vector \(\phi_i\) is the embedding produced by the model. This kernel matrix combines both the quality and diversity of the samples to guide the selection.

\paragraph{Subset Selection for DPP}
Given the kernel matrix \(L\), the probability of selecting a subset \(Y\) of samples is proportional to the determinant of the corresponding submatrix \(L_Y\):
\begin{equation}
    P_L(Y) \propto \text{det}(L_Y).
\end{equation}

The determinant measures how diverse the selected samples are—higher determinants indicate more diverse selections.

To efficiently select a subset, we use a greedy algorithm to approximate the optimal selection by iteratively choosing samples that maximize the determinant~\citep{ChenZZ18}.


\subsection{P3 Architecture}
The P3 framework integrates Self-Paced Learning (SPL) with Determinantal Point Processes (DPP) to dynamically adjust data pruning during training process. It increases the difficulty of training samples based on the model's learning capacity while maintaining diversity, which helps prevent overfitting and improve generalization.

In this stage-wise data selection approach, described in Algorithm~\ref{alg:stageWiseTraining}, SPL selects samples that match the model’s current capability, while DPP ensures the selected samples are diverse. This combined strategy optimizes the training data selection, enabling the model to gradually learn more challenging tasks effectively.

\begin{algorithm}[h]
\caption{P3 Architecture}
\begin{algorithmic}
\Require training dataset $D$, model $\theta$, number of epochs $E$, number of selected samples $k$, adjustment factor $\alpha$, self-paced factor $\lambda$
\Function{DynamicSelect}{$D$, $\theta$, $e$, $k$, $\alpha$, $\lambda$}\label{alg:stageWiseTraining}
    \State $d_e \gets \text{ComputeDifficulties}(\theta, D, e)$
    \State $S_{filtered} \gets \{d \in D \mid d_{\text{adj}, e} \leq \lambda(e)\}$
    \State $K \gets \text{ComputeKernelMatrix}(S_{filtered})$
    \State $indices \gets \text{DPP}(K, k)$
    \State $S_{e} \gets \{S_{filtered}[i] \mid i \in indices\}$
    \State \Return $S$
\EndFunction
\For{epoch $e = 1$ \textbf{to} $E$}
    \State $S_{e} \gets \text{DynamicSelect}(D, \theta, e, k, \alpha, \lambda)$
    \State $\theta, \text{loss} \gets \text{TrainModel}(\theta, S_{e})$
\EndFor
\end{algorithmic}
\end{algorithm}

The P3 architecture effectively balances sample difficulty and diversity throughout training. Self-paced learning adapts the difficulty threshold \(\lambda\) over epochs, allowing the model to gradually take on more challenging data as it learns. Meanwhile, DPP ensures that the training data remains diverse, which improves generalization and helps the model avoid overfitting.

The stage-wise training mechanism follows a straightforward flow: in each epoch, the most appropriate samples are selected based on their difficulty, and DPP is used to promote diversity among the selected samples. This combined approach results in an efficient and adaptive training process, enabling the model to better handle complex tasks over time.

\begin{table*}[t]
\small
\centering
\caption{Experimental results on the APPS dataset. The model trains on 1,500 samples per epoch from a 9,771-sample set over 5 epochs. GPTNeo, with fewer parameters, gets a 3,000-sample initial warm-up, not used for other models.\vspace{-0.3cm}}
\label{tab:APPS}
\begin{tabularx}{\linewidth}{ll|>{\centering\arraybackslash}X >{\centering\arraybackslash}X | >{\centering\arraybackslash}X >{\centering\arraybackslash}X | >{\centering\arraybackslash}X >{\centering\arraybackslash}X | >{\centering\arraybackslash}X >{\centering\arraybackslash}X}
\toprule
\multicolumn{2}{l}{} & \multicolumn{2}{c|}{\textbf{GPTNeo-2.7B}} & \multicolumn{2}{c|}{\textbf{Baichuan2-7B}} & \multicolumn{2}{c|}{\textbf{Llama3-8B}} & \multicolumn{2}{c}{\textbf{Mistral2-7B}} \\
\cmidrule(l){3-4} \cmidrule(l){5-6} \cmidrule(l){7-8} \cmidrule(l){9-10}
\multicolumn{2}{l}{\multirow{-2}{*}{Method}} & \textbf{ACC} & \textbf{BLEU} & \textbf{ACC} & \textbf{BLEU} & \textbf{ACC} & \textbf{BLEU} & \textbf{ACC} & \textbf{BLEU} \\
\midrule
\multirow{5}{*}{Baseline} & Without SFT       & 3.46\% & 27.41 & 0.29\% & 16.07 & 1.23\% & 16.84 & 1.42\% & 52.48 \\
                          & Random            & \underline{8.66\%} & \underline{52.37} & 9.70\% & 50.53 & 27.44\% & 66.98 & 25.31\% & 65.48 \\
                          & D2Pruning         & 5.79\%    & 39.36    & 9.83\%    & 45.91 & 28.52\% & 66.39 & 26.33\% & 64.96 \\
                          & RHO-Loss          & 8.52\%    & 47.50  & 12.81\%   & \textbf{56.82} & 32.18\% & 68.73 & \underline{30.63\%} & 67.33 \\
                          & Cherry            & 4.99\%    & 22.63    & 10.33\%   & 41.60  & 20.18\% & 49.95 & 20.14\% & 61.23 \\
\midrule
\multirow{3}{*}{CL} & Answer Row          & 6.47\% & 41.10 & 10.13\% & 43.32 & \underline{32.78\%} & \underline{69.01}  & 29.54\% & \textbf{67.82} \\
                                          & Answer Length  & 6.68\% & 34.78 & \underline{14.33\%} & 47.06 & 32.56\% & 67.43 & 26.06\% & 64.27 \\
                                          & Question Length& 8.62\% & 46.49 & 12.63\% & 52.77 & 32.04\% & \textbf{69.93} & 27.14\% & 65.83 \\
\midrule
Ours & P3 & \textbf{10.69\%} & \textbf{54.04} & \textbf{14.45\%} & \underline{56.59} & \textbf{33.09\%} & 68.06 & \textbf{31.04\%} & \underline{67.47} \\
\bottomrule
\end{tabularx}
\end{table*}

\section{Experiments}
In this section, we conduct extensive experiments to address the following research questions:
\vspace{-0.1cm}
\begin{itemize}[leftmargin=25pt,itemsep=1.5pt]
    \item [\textbf{RQ1}] How does our framework compare to other methods?
    \item [\textbf{RQ2}] How does P3 perform compared to training with the full dataset across different tasks and models?
    \item [\textbf{RQ3}] How does policy-based difficulty measurement compare to other metrics, such as model-based and manual assessments?
    \item [\textbf{RQ4}] How effectively does DPP contribute to the performance when integrated with SPL?
    \item [\textbf{RQ5}] How do the hyperparameters of P3 influence overall model performance?
\end{itemize}

\subsection{Experimental Setups}
\subsubsection{\textbf{Datasets}}
We validate our approach on two well-established LLM datasets: APPS~\citep{hendrycksapps2021} and MATH~\citep{hendrycksmath2021}, both focused on reasoning.

\vspace{5pt}\noindent\textbf{APPS}~\citep{hendrycksapps2021} is a benchmark for code generation in LLMs. It provides problem descriptions, Python solutions, test cases, function names, and metadata. We use the Introductory level tasks, which include about 3,650 training samples. Due to input length constraints, we randomly select 2-3 solutions per problem, resulting in 9,771 training samples, and evaluate the model on 1,000 test samples.

\vspace{5pt}\noindent\textbf{MATH}~\citep{hendrycksmath2021} contains high school math competition problems across multiple subjects, each with a detailed solution in LaTeX. Problems are rated from 1 (easiest) to 5 (hardest). We use subsets: Algebra (1,744 training, 1,187 test samples) and Counting \& Probability (771 training, 474 test samples).

\subsubsection{\textbf{Models}} 
To compare model performance, we fine-tune three LLMs of different sizes: \textbf{GPTNeo-2.7B}~\citep{gpt-neo}, \textbf{Baichuan2-7B-chat}~\citep{baichuan2023baichuan2}, \textbf{Llama3-8B-instruct}~\citep{llama3modelcard}, and \textbf{Mistral-7B-Instruct-v0.2}~\citep{jiang2023mistral7b} on the APPS and MATH datasets. For MATH, we excluded GPTNeo-2.7B because, despite being trained on GitHub code data, its smaller parameter size makes it unsuitable for the complex reasoning required in MATH. 

\begin{table*}[t]
  \centering
    \caption{Experimental results on the MATH dataset. For Algebra, the model is trained on 300 out of a total of 1,744 samples over 5 epochs, with testing on 1,187 samples. For Counting and Probability, the model trains on 150 out of 771 samples per epoch, testing on 474 samples. No warm-up is used.\vspace{-0.3cm}}
  \label{tab:MATH}
  \small  
  \begin{tabular}{ll|ccc|ccc}
    \toprule
    \multicolumn{2}{l}{} & \multicolumn{3}{c}{\textbf{Algebra}} & \multicolumn{3}{c}{\textbf{Counting and Probability}} \\
    \cmidrule(l){3-5} \cmidrule(l){6-8}
    \multicolumn{2}{l}{\multirow{-2}{*}{Method}} & \textbf{Baichuan2-7B} & \textbf{Llama3-8B} & \textbf{Mistral-7B} & \textbf{Baichuan2-7B} & \textbf{Llama3-8B} & \textbf{Mistral-7B} \\
    \midrule
    \multirow{5}{*}{Baseline} & Without SFT       & 2.30\% & 6.07\% & 6.49\% & 1.05\% & 5.48\% & 2.74\% \\
                              & Random            & 6.32\% & 11.7\% & 5.56\% & \underline{5.49\%} & 8.44\% & 6.33\%\\
                              & D2Pruning         & 4.30\%    & 16.01\%  & 6.74\%  & 2.53\%    & 12.23\% & 4.64\%\\
                              & RHO-Loss          & 6.40\%    & \textbf{19.04\%} & 6.74\% & 4.43\%   & \underline{13.08\%} & \underline{7.38\% }\\
                              & Cherry            & 3.54\% & 11.8\% & 6.23\% & 2.53\%    & 6.96\%  & 4.22\% \\

    \midrule
    \multirow{3}{*}{CL} & Level          & 6.57\% & 15.2\% & \underline{8.34\%} & 4.01\% & 9.49\% & 5.06\%\\
                                              & Answer Length  & 5.39\% & 13.7\% & 7.58\% & 3.38\% & 7.38\% & 1.89\%\\
                                              & Question Length& \underline{8.00\%} & 11.6\% & 7.58\% & 4.01\% & 8.23\% & 5.91\% \\
    \midrule
    Ours & P3 & \textbf{9.35\%} & \underline{17.10\%} & \textbf{9.77\%} & \textbf{7.81\%} &  \textbf{13.29\%} & \textbf{8.23\%}\\
    \bottomrule
  \end{tabular}
\end{table*}

\vspace{-0.1cm}
\subsubsection{\textbf{Evaluation Metrics}}
For APPS, we assess code generation using BLEU scores and test case accuracy. For MATH, model success is based on final answer accuracy from step-by-step LaTeX solutions. In all metrics, higher scores indicate better performance.

\vspace{-0.1cm}
\subsubsection{\textbf{Implementation Details}}
We use consistent parameters during fine-tuning. The initial learning rate is 0.0004, with a weight decay of 0.0001. To improve hardware efficiency, we use 4-bit quantization and set the computational data type to torch.float16. We apply Low-Rank Adaptation (LoRA) with \(r=64\), \(\text{lora\_alpha}=16\), and a 0.1 dropout rate for the LoRA layer. APPS experiments are conducted on A6000 48G, while MATH experiments use A100 80G.

\begin{table*}[t]
\small
  \centering
  \caption{Ablation studies, comparing the results of using DPP for diverse data selection versus not using DPP.\vspace{-0.3cm}}
  \label{tab:ablation}
  \small
  \begin{tabularx}{\textwidth}{l|>{\centering\arraybackslash}X>{\centering\arraybackslash}X>{\centering\arraybackslash}X>{\centering\arraybackslash}X|>{\centering\arraybackslash}X>{\centering\arraybackslash}X>
  {\centering\arraybackslash}X|>{\centering\arraybackslash}X>
  {\centering\arraybackslash}X>{\centering\arraybackslash}X}   
    \toprule
    & \multicolumn{4}{c|}{\textbf{APPS}} & \multicolumn{3}{c|}{\textbf{Algebra}} & \multicolumn{3}{c}{\textbf{Counting and Probability}} \\
    \cmidrule(r){2-5} \cmidrule(lr){6-8} \cmidrule(l){9-11}
    & \textbf{GPTNeo} & \textbf{Baichuan2} & \textbf{Llama3} & \textbf{Mistral2} & \textbf{Baichuan2} & \textbf{Llama3} & \textbf{Mistral2}& \textbf{Baichuan2} & \textbf{Llama3} & \textbf{Mistral2}\\
    \midrule
    SPL & 8.71\% & 11.97\% & 32.85\% & 29.25\% & 9.10\% & 15.42\% & 9.10\% & 5.06\% & 8.86\% & 5.91\%\\
    P3 & 10.69\% & 14.45\% & 33.09\% & 31.04\% & 9.35\% & 17.10\% & 9.77\% & 7.81\% & 13.29\% & 8.23\%\\
    \bottomrule
  \end{tabularx}
\end{table*}

\begin{figure*}[ht]
    \centering
    \includegraphics[width=\linewidth]{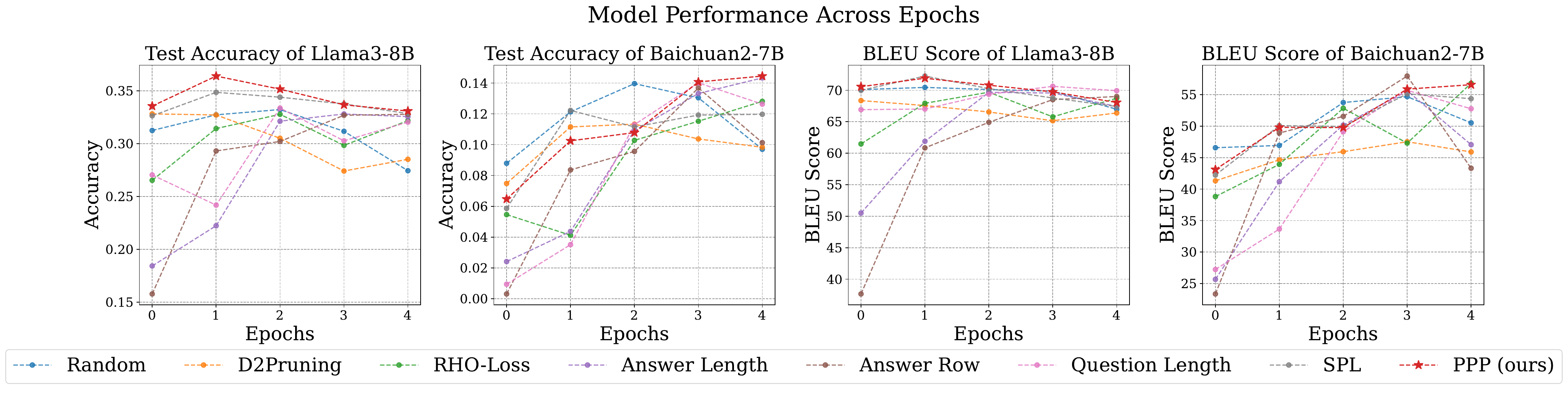}
    \vspace{-0.6cm}
    \caption{APPS dataset comparison over 5 epochs. Test accuracy and BLEU score show that our P3 method outperforms other baselines with faster learning and stronger results.}
    \label{fig:1_4}
\end{figure*}

\subsection{Baselines}
We compare our method against several baselines in two categories: state-of-the-art coreset selection techniques~\citep{pmlr-v162-mindermann22a,maharana2023d2} and advanced data selection methods in NLP~\citep{li2024quantityqualityboostingllm}, as well as curriculum learning approaches that determine training sample difficulty.

The baseline methods are:

\begin{itemize}[leftmargin=*,itemsep=1pt] \setlength{\partopsep}{0pt} 
\vspace{-0.2cm}
\item \textbf{D2Pruning}~\citep{maharana2023d2}: Represents the dataset as a graph and selects coresets using message passing, updating difficulty scores based on neighboring examples.

\item \textbf{RHO-Loss}~\citep{pmlr-v162-mindermann22a}: Estimates the impact of each data point on reducing loss for unseen data, focusing on non-noisy, non-redundant, and task-relevant examples.

\item \textbf{Cherry}~\citep{li2024quantityqualityboostingllm}: Uses the Instruction-Following Difficulty (IFD) metric to select high-quality samples from open-source datasets.

\item \textbf{Curriculum Learning}: Involves sorting training data by predefined difficulty and training sequentially, starting with easy examples. For APPS, we use the number of code lines and tokens; for MATH, we use manually assigned difficulty levels and token counts.
\end{itemize}

\subsection{Experimental Results}
\subsubsection{\textbf{Evaluation on APPS}}
Table~\ref{tab:APPS} presents the evaluation of different optimization strategies on the APPS dataset using three language models. Baselines like D2Pruning and RHO-Loss yield varied results in Test Case Accuracy and BLEU scores. Random performs well with GPTNeo-2.7B, while RHO-Loss shows consistently strong performance across all models. Curriculum learning approaches using metrics like Answer Row, Answer Length, and Question Length showed mixed improvements.

Figure~\ref{fig:1_4} compares methods over five epochs in terms of accuracy and BLEU. The P3 method demonstrates rapid learning in early epochs and consistently outperforms baselines, with lower overfitting on Llama3-8B compared to Random and D2Pruning.

\subsubsection{\textbf{Evaluation on MATH}}
Table~\ref{tab:MATH} shows results for the MATH dataset on Algebra and Counting \& Probability tasks using Baichuan2-7B-chat and Llama3-8B-instruct. For Algebra, P3 achieves the highest accuracy of 9.35\% on Baichuan2-7B-chat and 15.4\% on Llama3-8B-instruct, comparable to the Level-based method. In Counting \& Probability, P3 leads with 11.18\% accuracy on Llama3-8B-instruct, underscoring its effectiveness for mathematical tasks and optimizing model performance.

\subsubsection{\textbf{In-Depth Analysis}}
In this section, we compare our framework to existing baselines (RQ1) and assess model performance.

\vspace{3pt}\noindent\textbf{Answer to RQ1:}
\vspace{-0.2cm}
\begin{itemize}[leftmargin=*, itemsep=1.5pt]
    \item On the APPS dataset, our method outperforms RHO-Loss in both Test Case Accuracy and BLEU. Specifically, Test Case Accuracy improves by 25.5\% on GPTNeo-2.7B, 12.8\% on Baichuan2-7B-chat, and 2.83\% on Llama3-8b-ins, consistently achieving better results throughout training.
    \item On MATH, P3 shows superior performance for both Algebra and Counting \& Probability tasks.
\end{itemize}
\vspace{-0.1cm}


\subsubsection{\textbf{Comparison with Full dataset}}
In this section, we compare P3 to full dataset training across tasks and models.
\vspace{-0.1cm}

\begin{table}[h!]
\small
\centering
\caption{APPS: Full vs. P3 performance.\vspace{-0.3cm}}
\begin{tabularx}{1\linewidth}{ll|c|>{\centering\arraybackslash}X}
\toprule
\textbf{APPS} &  & \textbf{Test Acc} & \textbf{BLEU} \\
\midrule
\textbf{GPTNeo} & \textbf{Full (9771*5)} & 0.0799 & 45.17 \\
                 & \textbf{P3 (1500*5)} & 0.1069 & 54.04 \\
\midrule
\textbf{Baichuan2} & \textbf{Full (9771*5)} & 0.1681 & 57.89 \\
                      & \textbf{P3 (1500*5)} & 0.1445 & 56.59 \\
\midrule
\textbf{Llama3} & \textbf{Full (9771*5)} & 0.3096 & 65.02 \\
                & \textbf{P3 (1500*5)} & 0.3309 & 68.06 \\
\midrule
\textbf{Mistral2} & \textbf{Full (9771*5)} & 0.2761 & 67.01 \\
                & \textbf{P3 (1500*5)} & 0.3104 & 67.47 \\
\bottomrule
\end{tabularx}

\label{tab:apps_performance}
\end{table}
\vspace{-0.3cm}

\begin{table}[h!]
\small
\centering
\caption{MATH: Full vs. P3 performance.\vspace{-0.3cm}} 
\begin{tabularx}{\linewidth}{l@{\hskip 4pt}l@{\hskip 2pt}|c@{\hskip 2pt}c|c@{\hskip 2pt}c}
\toprule
\textbf{MATH} &  & \multicolumn{2}{c|}{\textbf{Algebra}} & \multicolumn{2}{c}{\textbf{C\&P}} \\
\cmidrule{3-6}
& & \textbf{\scriptsize{Full (1744*5)}} & \textbf{\scriptsize{ P3 (300*5)}} & \textbf{\scriptsize{Full (771*5)}} & \textbf{\scriptsize{ P3 (150*5)}} \\
\midrule
\textbf{Baichuan2} &  & 0.0851 & 0.0935 & 0.0675 & 0.0781 \\
\midrule
\textbf{Llama3} &  & 0.1592 & 0.1710 & 0.0992 & 0.1329 \\
\midrule
\textbf{Mistral2} &  & 0.0834 & 0.0977 & 0.0780 & 0.0823\\
\bottomrule
\end{tabularx}
\label{tab:math_performance}
\end{table}
\vspace{-0.1cm}

\vspace{3pt}\noindent\textbf{Answer to RQ2:}
\vspace{-0.2cm}
\begin{itemize}[leftmargin=*]
    \item Tables~\ref{tab:apps_performance} and \ref{tab:math_performance} show that P3, using fewer samples per epoch, achieves results comparable to full dataset fine-tuning, especially in BLEU scores on APPS. On MATH, P3 performs well in Counting\_and\_Probability and matches full dataset training in Algebra, demonstrating that P3 can achieve similar performance to full fine-tuning with significantly reduced data.
\end{itemize}

\subsubsection{\textbf{Ablation Study}}
In this section, we examine the effects of the two core components of our data selection framework: difficulty and diversity.

\vspace{5pt}\noindent\textbf{Answer to RQ3:} 
We analyze the impact of different difficulty metrics on the APPS and MATH datasets:
\vspace{-0.2cm}
\begin{itemize}[leftmargin=10pt,itemsep=1.5pt]
    \item \textbf{Manual Metrics}: Metrics based on human annotations (level or answer rows) show limited effectiveness in guiding the model's learning process.
    \item \textbf{Model-Based Metrics}: Metrics using token counts ( question/answer length) perform slightly better but fail to fully capture task complexity.
    \item \textbf{Policy-Based Metrics (P3)}: Policy-driven metrics dynamically assess task difficulty based on model actions. As shown in Table~\ref{tab:ablation}, P3 achieves superior results, with a Test Case Accuracy of 33.09\% on APPS and 13.29\% on MATH using Llama3-8b-ins. This suggests that policy-based metrics effectively capture task complexity and adapt well to the model’s learning, outperforming manual and model-based approaches.
\end{itemize}
\vspace{-0.2cm}
\vspace{5pt}\noindent\textbf{Answer to RQ4:} 
\vspace{-0.2cm}
\begin{itemize}[leftmargin=*]
    \item \textbf{Impact of Diversity (DPP)}: The t-SNE visualization in Figure~\ref{fig:analysis} shows diverse and uniformly distributed data across epochs, indicating effective data variety. Table~\ref{tab:ablation} confirms that incorporating DPP improves model accuracy over SPL on both APPS and MATH datasets, validating DPP’s role in enhancing performance through diverse data selection.
\end{itemize}

\subsubsection{\textbf{Hyper-Parameter Analysis}}
In this section, we analyze how the number of selected samples and epochs affect the performance of the Baichuan-7B-chat model on the Algebra dataset.
\vspace{-0.2cm}

\begin{figure}[h!]
    \centering
    \includegraphics[width=\linewidth]{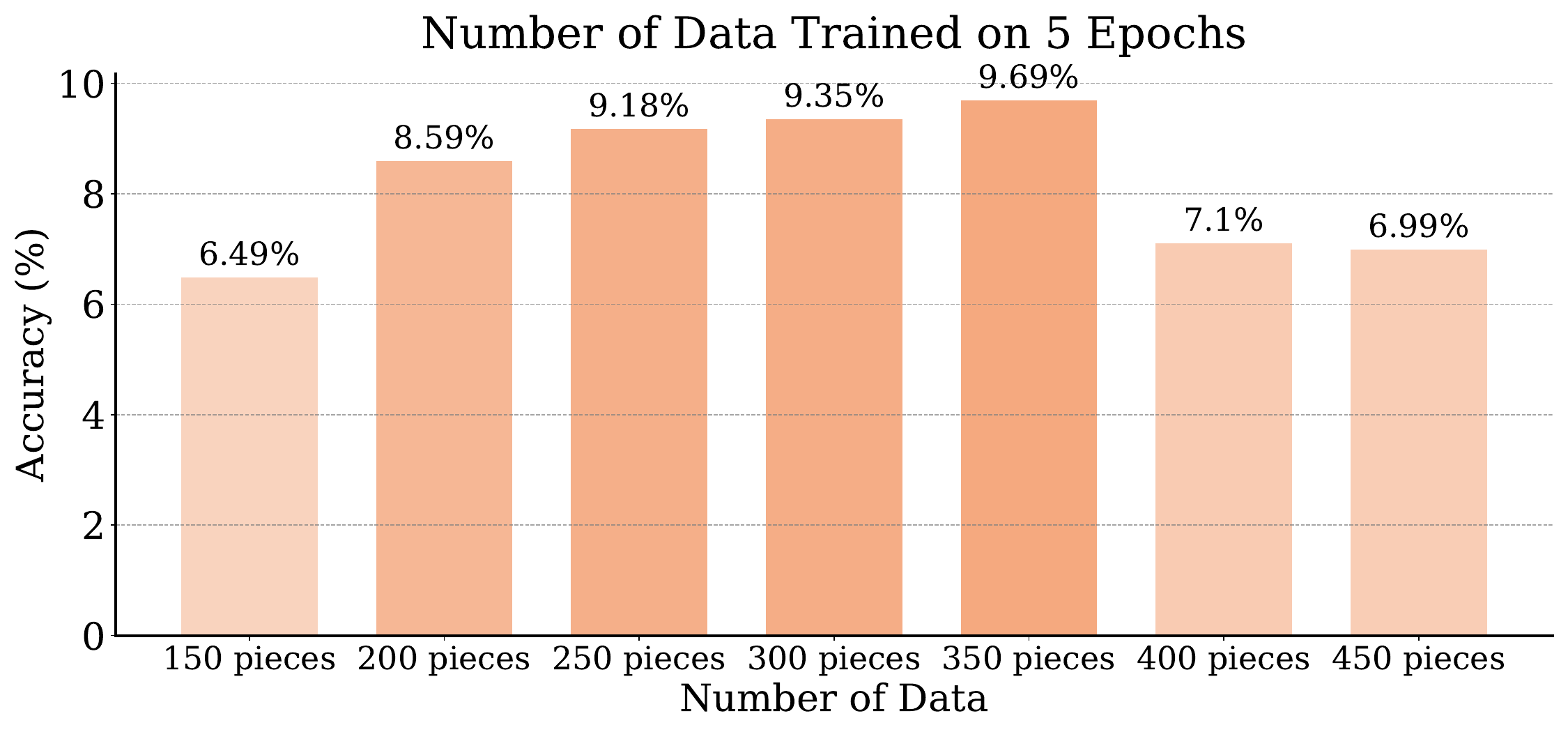}
    \vspace{-0.8cm}
    \caption{Effect of data size on acc over 5 epochs.}
    \label{fig:case1}
\end{figure}
\vspace{-0.5cm}
\begin{figure}[h!]
    \centering
    \includegraphics[width=\linewidth]{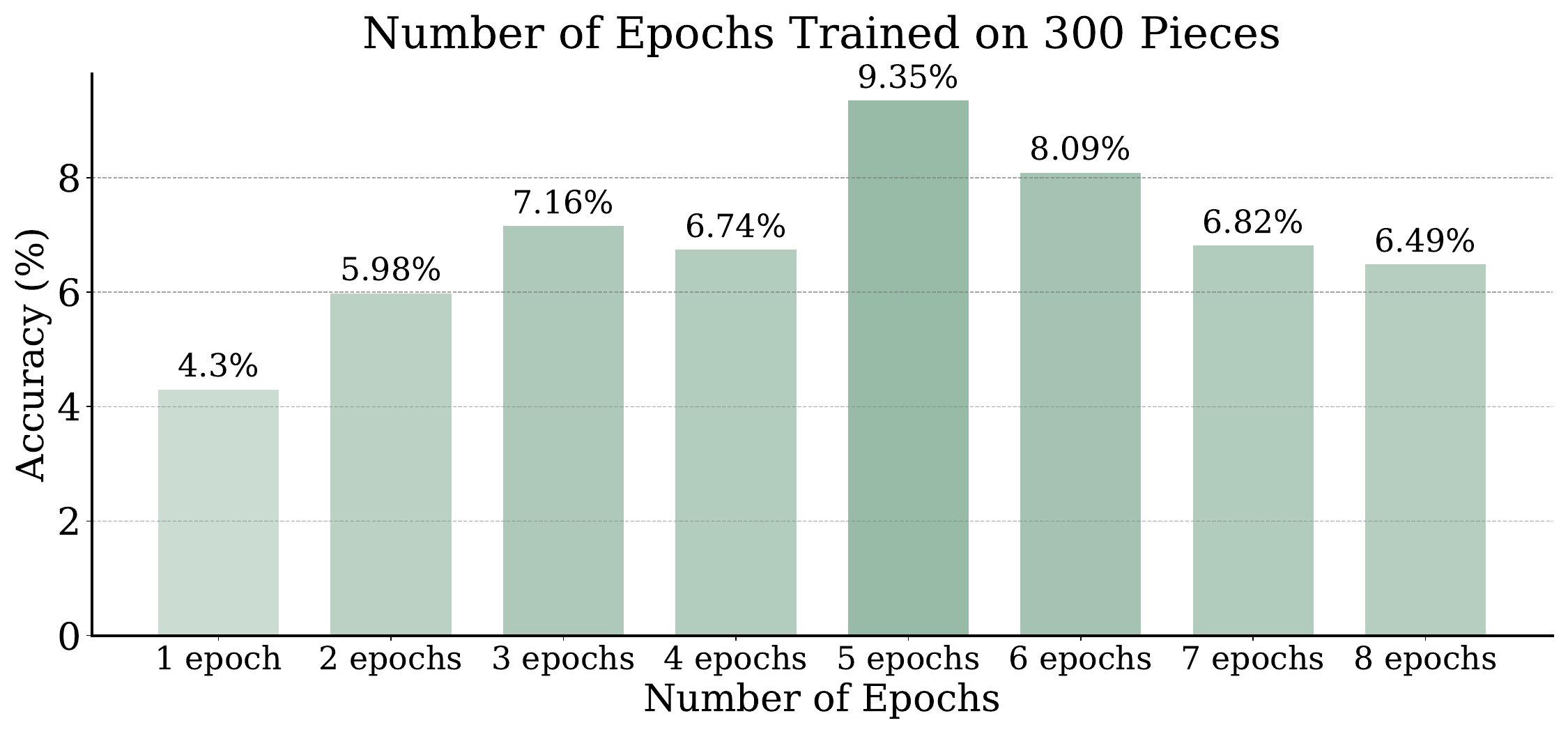}
    \vspace{-0.8cm}
    \caption{Effect of epochs on acc with 300 samples.}
    \label{fig:case2}
\end{figure}

\vspace{-0.4cm}

\vspace{5pt}\noindent\textbf{Answer to RQ5:}
\vspace{-0.2cm}
\begin{itemize}[leftmargin=*, itemsep=1.5pt]
    \item Figure~\ref{fig:case1} shows that accuracy improves as more training data is added, but gains diminish beyond a certain point, indicating limited benefit from excessive data.
    \item Figure~\ref{fig:case2} demonstrates that performance initially improves with more epochs but declines after a certain point, suggesting an optimal stopping point is needed to avoid overfitting.
\end{itemize}

\section{Conclusion}
In this paper, we present P3, a novel framework for data pruning during training to fine-tune LLM. Experiments on APPS and MATH datasets demonstrate that P3 outperforms existing methods, offering an adaptive strategy for optimizing LLM performance. This work highlights the importance of adaptive data selection for efficient training and robust model performance.

\end{document}